# A Unified Physics-Aware Quantum Machine Learning Framework across Power GaN HEMTs and Logic Nanowire FETs: Predicting Unseen Process Splits and Held-Out Geometry Combinations with Lower Error and Tighter Split-to-Split Variability

Rushat Rai[1], Yun-Yuan Wang[2], Autsada Kakaen[3], Pei-Jie Chang[1], Doan Viet Nguyen[1], Yuan-Chieh Chiu[1], Doldet Tantraviwat[3], Niall Tumilty[1], Simon See[4], Wen-Jay Lee[5,*], Tai-Yue Li[5,*], Nan-Yow Chen[5,*], and Tian-Li Wu[1,*]

[1] National Yang Ming Chiao Tung University, Taiwan. [2] NVIDIA AI Technology Center, Taiwan. [3] Chiang Mai University, Thailand. [4] NVIDIA AI Technology Center, Singapore. [5] National Center for High-performance Computing, Taiwan.

*E-mail: {wjlee, 2503001, nanyow}@niar.org.tw, tlwu@nycu.edu.tw

***Abstract-*** We present a unified reinforcement-learning (RL) framework that discovers compact parametrized quantum circuits (PQCs) for data-scarce device modeling. A graph neural network (GNN) policy optimized by proximal policy optimization (PPO) searches circuit architectures using leave-one-group-out cross-validation (LOGOCV) error on held-out process or geometry groups as the reward. The framework achieves the lowest mean absolute error (MAE) on all 11 targets versus six classical baselines, with 59% lower error ($I_{off}$) and 81% tighter fold variability ($V_{TH}$) for HEMTs and 84% lower error ($V_{TH}$, SS, $I_{off}$) and 82% tighter fold variability ($I_{off}$) for NWFETs. These results demonstrate the potential of RL-selected, classically simulated PQCs as compact surrogates with low OOD error and improved physical consistency, despite imposing no explicit physical constraints, penalty terms, or device-specific equations, on the two evaluated device datasets.

## I. INTRODUCTION

Advanced device architectures are evolving to meet the distinct demands of high-voltage power conversion and sub-3-nm logic. p-GaN gate GaN HEMTs enable normally-off operation with controllable threshold voltage for high-performance power switching [1], while gate-all-around nanowire FETs (NWFETs) suppress short-channel effects and off-state leakage for continued scaling toward low-power logic [2].

Their optimization is limited by costly fabrication and TCAD cycles and by stochastic variation from charge trapping, thermal gradients, and parameter extraction. High-value design exploration thus requires physically consistent extrapolation to unfabricated process splits and held-out geometry combinations, not merely interpolation among measured devices (**Fig. 1**).

Existing surrogates often embed device-specific equations in the architecture [16]-[18], impeding reuse across device families, or report only within-distribution tests [16][17]. We instead impose no explicit physical constraints, penalty terms, or device-specific equations, making out-of-distribution (OOD) prediction the explicit objective; physical consistency emerges from the OOD reward. Parametrized quantum circuits (PQCs) provide a compact Fourier-structured function class whose expressivity grows with depth and data re-uploading [7][8] rather than parameter count [9]. A relational graph-attention policy represents candidate circuits as heterogeneous graphs, optimized by PPO with a leave-one-group-out cross-validation reward, selecting architectures for unseen process or geometry groups rather than average training fit.

A single search framework applies to both planar HEMTs and 3D NWFETs without re-deriving device equations, discovering a circuit optimized for each (**Fig. 4**): HEMTs use physics-based pretraining followed by calibration to measured wafer data, while NWFETs learn directly from 3D TCAD. Across 11 targets, the selected circuits reduce OOD error by 6-34% for HEMTs and 25-39% for NWFETs, with about 5× tighter HEMT split-to-split variability. To our knowledge, this is the first experimentally validated, RL architecture-searched PQC framework [10] for multi-target device modeling optimized directly for OOD prediction (**Table 3**).

## II. METHODOLOGY

**Dataset preparation:** To simulate p-GaN HEMTs for pretraining, a 1D electrostatic solver incorporating Mg-acceptor ionization, polarization charge, 2D electron-gas (2DEG) formation, and interface traps [3]–[5] (**Fig. 5**) was developed. Solved by finite-volume discretization with damped Newton iteration, it extracts full transfer curves and six targets: forward/reverse threshold voltages ($V_{TH,for}$/$V_{TH,rev}$), hysteresis ($\Delta V_{TH}$), subthreshold swing (ss), on-current ($I_{ON}$), and off-current ($I_{OFF}$). Electrostatic analysis confirms physical validity (**Fig. 3**): smooth band bending, gate-modulated surface potential, enhancement-mode operation with expected 2DEG densities, and charge conservation at the sweep turnaround. The solver generates 1,785 devices/h on a CPU (~90× faster than TCAD), while a PhysicsNeMo PINN reaches 12,898 devices/h on a GPU (~650× faster) [6] (**Table 3**) to augment it. The experimental HEMT dataset comprises 544 devices spanning unique process splits and wafer positions; the NWFET dataset comprises 1,914 3D-TCAD devices for five targets across eight groups defined by channel length, nanowire radius, and dielectric thickness.

**Encoding and circuit search:** Device features map one-to-one to qubits by angle encoding: five process/position inputs for HEMTs and three geometry inputs (channel length ($L_g$), radius ($r$), dielectric thickness ($t$)) for NWFETs. The agent selects a pre-encoder, $R_X$, $R_Y$, or $R_Z$ rotations, and entanglers from CNOT, CRX, CRY, and CRZ families arranged as chain, ring, all-to-all, or even/odd pairs. The shared library contains 5 rotation and 24 entangling blocks plus data re-uploading, with candidate depth 3-12 blocks and at most four re-uploads. Validity masks require at least one rotation and one entangler, forbid consecutive re-uploads, and allow early termination, limiting unproductive exploration.

**Readout:** Single-qubit Z expectations pass through a trainable tempered tanh and linear map to standardized multi-target outputs. For measured HEMTs, a wafer-position correction is added to the quantum prediction, letting the PQC capture process-dependent behavior while the small residual head accounts for spatial drift across the wafer.

**GNN policy and RL:** Each partial circuit is encoded as a typed graph with feature, qubit, operation-block, and readout nodes linked by circuit-specific relations (**Fig. 2**). A relational graph attention network serves as a shared actor-critic backbone [11], pooling masked node embeddings with global circuit features. The actor selects the pre-encoder, feature encoding, next valid block, or termination; the critic estimates terminal reward. PPO with generalized advantage estimation updates the policy [12][13], and a 0.001 per-block penalty discourages unnecessary depth. This preserves connectivity across variable-length candidates and device-specific qubit counts.

**Training and selection.** HEMT training uses three stages: synthetic electrostatic pretraining, wafer-position-head calibration with the PQC frozen, and joint experimental fine-tuning; NWFETs use direct single-stage TCAD training. Architecture selection follows nested LOGOCV: the inner reward uses the worst validation-group error with complexity and invalid-structure penalties, so a strong average cannot hide failure on one split; the selected candidate is evaluated on an untouched outer group. Hybrid quantum-classical training minimizes mean-squared error alone, with no physics-informed loss or constraint terms, using parameter-shift gradients [14] on NVIDIA CUDA-Q [15].

## III. RESULTS AND DISCUSSION

We compare our RL-selected hybrid quantum neural network (PQC) against six classical baselines on two OOD challenges: measured p-GaN HEMT process variation and 3D NWFET geometry scaling (**Fig. 7**). All models use the same group-wise validation protocol to prevent data leakage; results are averaged over 5 random seeds.

To isolate the model-class effect, the ANN baseline used the identical data pipeline, wafer-position correction, training stages, and LOGOCV splits as the PQC. It was swept across shallow-to-deep capacities plus a Fourier-feature ablation, and we report its best variant.

**Benchmark accuracy and efficiency:** The PQC has the lowest OOD MAE on all 11 targets versus the six classical baselines. For HEMTs, $V_{TH,for}$ MAE is 18.6% below the identically trained ANN and 48.0% below the worst baseline (PLS); $V_{TH,rev}$ MAE is 25.2% below the ANN and 42.1% below the worst baseline; and $I_{OFF}$ MAE is 23.5% below the ANN and 59% below the worst baseline (Elastic Net). $\Delta V_{TH}$, SS, and $I_{ON}$ MAE also improve, while forward threshold-voltage fold variability is 53% tighter than the ANN and 81% tighter (5.3×) than the worst baseline (Decision Tree). For NWFETs, the PQC leads on all five targets, $V_{TH}$, SS, and $I_{OFF}$ MAE are 21.4%, 29.7%, and 37.5% below the ANN, respectively, and about 84% below the worst baseline (PLS); $I_{OFF}$ fold variability is 82% tighter than the worst baseline (**Fig. 13**). It uses 2,460 trainable model parameters (4.7× fewer than the ANN) and sits on the best error-parameter Pareto frontier (**Fig. 8**).

**Physical consistency in HEMTs:** Although PQC and ANN have similar aggregate error distributions, the PQC better preserves threshold-hysteresis closure (**Fig. 9**), even though no hysteresis-closure constraint or physics-based penalty is applied during training, so this consistency arises from the OOD-rewarded architecture search rather than explicit enforcement. Consistency metrics pool out-of-fold predictions from models refitted per outer group. We measure the absolute deviation of ($V_{th,rev}$ - $V_{th,forw}$) from the independently predicted hysteresis ($\Delta V_{TH}$), which should be zero for consistent outputs. The PQC median consistency error is about 5.4× lower than the ANN's; the ANN's inconsistent tail reaches roughly 4× its own median (about 22× the PQC median). Residual decomposition shows recipe-level mean bias falls to about 0.12 interquartile range (IQR) versus about 0.4 IQR for ANN, whereas die-to-die pattern error stays similar at about 0.4-0.5 IQR. The improvement mainly reflects suppressed wafer-to-wafer drift without losing local variation.

**Retention of pretrained physics after fine-tuning:** To test whether calibration preserves the process behavior acquired during solver pretraining, the fine-tuned process cores were reevaluated on the synthetic corpus with the wafer-position residual head excluded (**Fig. 11**). Before fine-tuning, both models retained all 16 evaluable solver-trend signs, with mean Spearman agreement $\rho_s$ = 0.949 (ANN) and 0.935 (PQC). After fine-tuning, ANN sign retention fell to 75% and mean agreement to $\rho_s$ = 0.501, reversing all four $I_{on}$ trends, whereas PQC retained all 16 signs with mean $\rho_s$ = 0.875. PQC also exceeded ANN in sign retention and trend agreement across all LOGO folds, with substantially lower fine-tuning-induced degradation.

**Joint-geometry OOD generalization in NWFETs:** To evaluate electrostatic gate control across subthreshold and on-state regimes, we model three metrics: off-current, subthreshold swing, and maximum transconductance. For these radius-sensitive targets, the PQC reduces out-of-fold MAE relative to the ANN by 50.1%, 46.3%, and 39.2%, respectively. Because weak-inversion current is exponentially sensitive to small potential-barrier errors in held-out geometries, we bin normalized MAE by distance to the nearest training geometry (**Fig. 10**). Both models degrade deeper into OOD space, but the PQC degrades more slowly: its advantage grows from 45.5% ($I_{OFF}$) and 30.0% (SS) in the nearest bin to 51.6% and 58.9% in the deepest bin. The widening margin indicates the quantum prior captures geometry-dependent electrostatic scaling rather than memorizing sampled points. To verify that OOD gains reflect genuine physical consistency, we evaluate preservation of geometry-dependent TCAD behavior (**Fig. 12**). Along controlled slices with two geometry variables fixed, both models reproduce the dominant trends, but only the PQC tracks highly nonlinear responses such as the non-monotonic $g_{max}$ dependence on dielectric thickness ($t$). Across all exact slices, the PQC achieves higher directional agreement with TCAD than the ANN (0.928 vs. 0.903) and a finite-difference slope ratio much closer to unity (0.979 vs. 0.877). The PQC also yields lower normalized MAE in 30 of 32 target–geometry combinations, with a median relative reduction of 42.2% and a worst-octant value of 0.133 vs. 0.333. Because these folds withhold joint low/high combinations of $L_g$, $r$, and $t$ while using individual axis values seen elsewhere in training, the framework generalizes robustly to unseen joint geometries and missing design-space corners.

## IV. CONCLUSION

We demonstrate the first experimentally validated, RL-searched PQC framework optimized directly for OOD device prediction. **Table 4** highlights that, for the first time, the proposed framework uniquely combines multi-target OOD prediction, RL-searched quantum circuits, and scalability across distinct power and logic device families without embedding device-specific equations**.** A single physics-aware workflow generalizes across measured p-GaN HEMT process splits and unseen 3D-NWFET geometry combinations, achieving the lowest OOD MAE for all 11 targets among six classical baselines. Compared with the identically trained ANN, the PQC reduces MAE by up to 50.1% while using 4.7× fewer parameters; versus all baselines, reductions reach 59% for HEMTs and 84% for NWFETs. Crucially, it preserves all 16 pretrained HEMT physical trends, improves threshold–hysteresis consistency by 5.4×, and maintains nonlinear NWFET geometry scaling without explicit physical constraints. These results establish OOD-rewarded quantum architecture search as a framework to discover parameter-efficient surrogate for data-scarce device optimization.

**Acknowledgments:** Supported by 1) NVIDIA Academic Grant Program and Dr. Chi-Cheng Fu (NVIDIA), 2) Advanced Semiconductor Technology Research Center under Higher Education Sprout Project of Ministry of Education, Taiwan, and 3) NSTC 114-2223-E-A49-003-MY4, 114-2119-M-007-013, 114-222-E-492-002-MY2. 

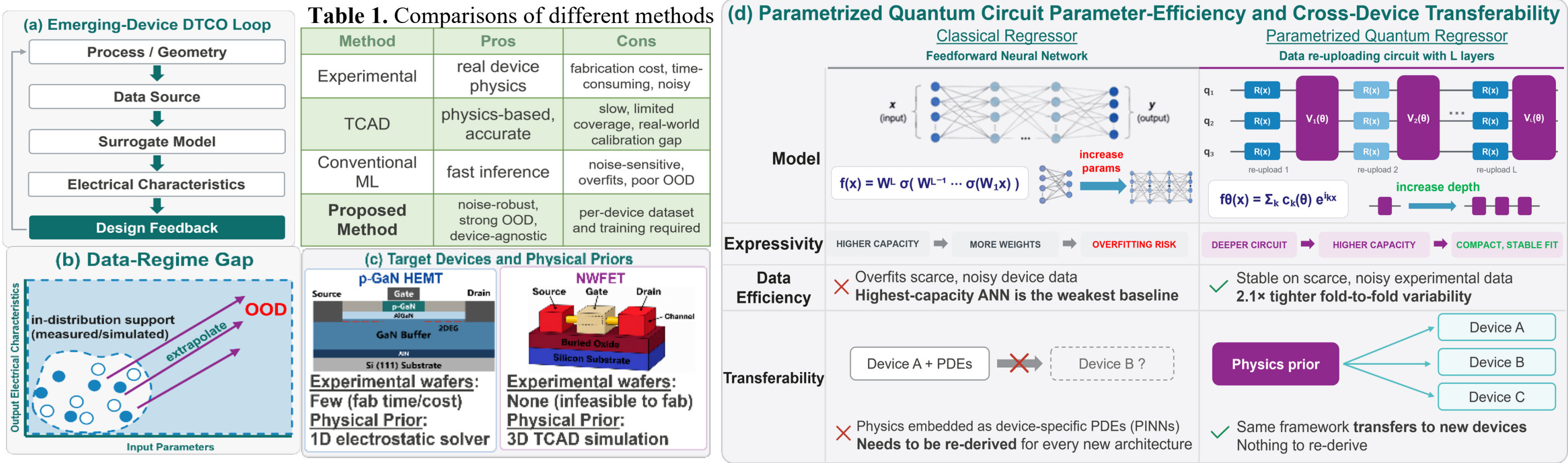

**Table 1.** Comparisons of different methods

| Method | Pros | Cons |
|---|---|---|
| Experimental | real device physics | fabrication cost, time-consuming, noisy |
| TCAD | physics-based, accurate | slow, limited coverage, real-world calibration gap |
| Conventional ML | fast inference | noise-sensitive, overfits, poor OOD |
| **Proposed Method** | noise-robust, strong OOD, device-agnostic | per-device dataset and training required |

**Fig. 1.** Emerging-device surrogate-modeling DTCO challenges and the proposed unified OOD modeling framework: (a) conventional DTCO workflow and modeling methods (Table 1); (b) high-value optimization requires physically consistent OOD extrapolation, where CML degrades; (c) target devices and data sources; and (d) data re-uploading PQCs provide parameter-efficient expressivity scaling and enable a common framework that adapts across devices without re-deriving PDEs.

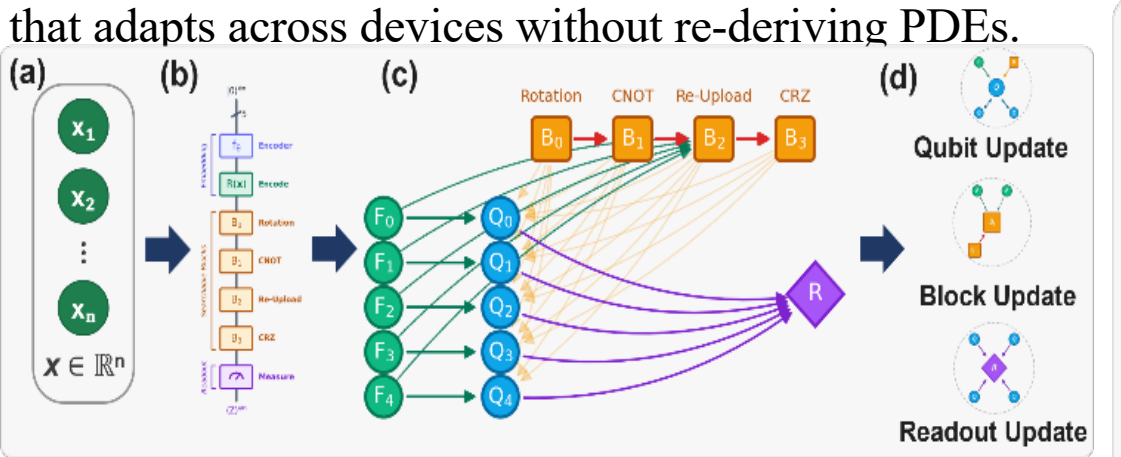

**Fig. 2.** RL-search framework for PQC architecture discovery. (a) Device features form the input vector. (b) Inputs are encoded into an n-qubit PQC with searchable rotation and entanglement blocks. (c) The PQC is mapped to a heterogeneous graph of feature, qubit, block, and readout nodes. (d) A relational graph-attention network (GNN) predicts the next circuit operation during RL search.

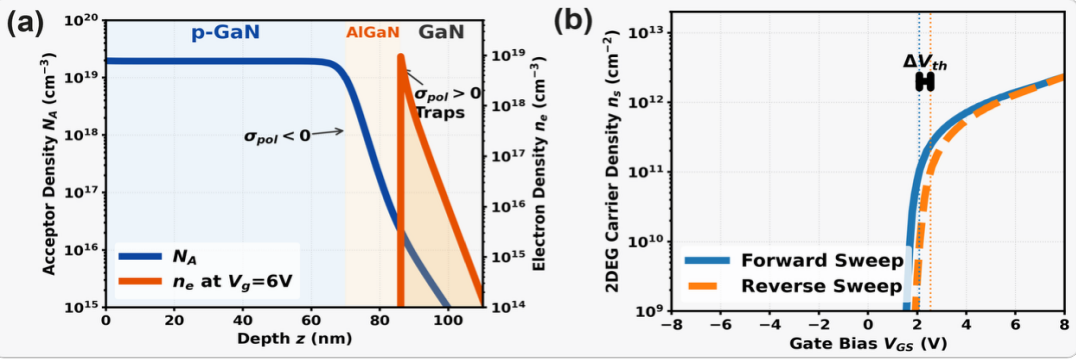

**Fig. 3.** Electrostatic validation of the p-GaN HEMT solver: (a) doping profile, and (b) simulated 2DEG density versus gate bias.

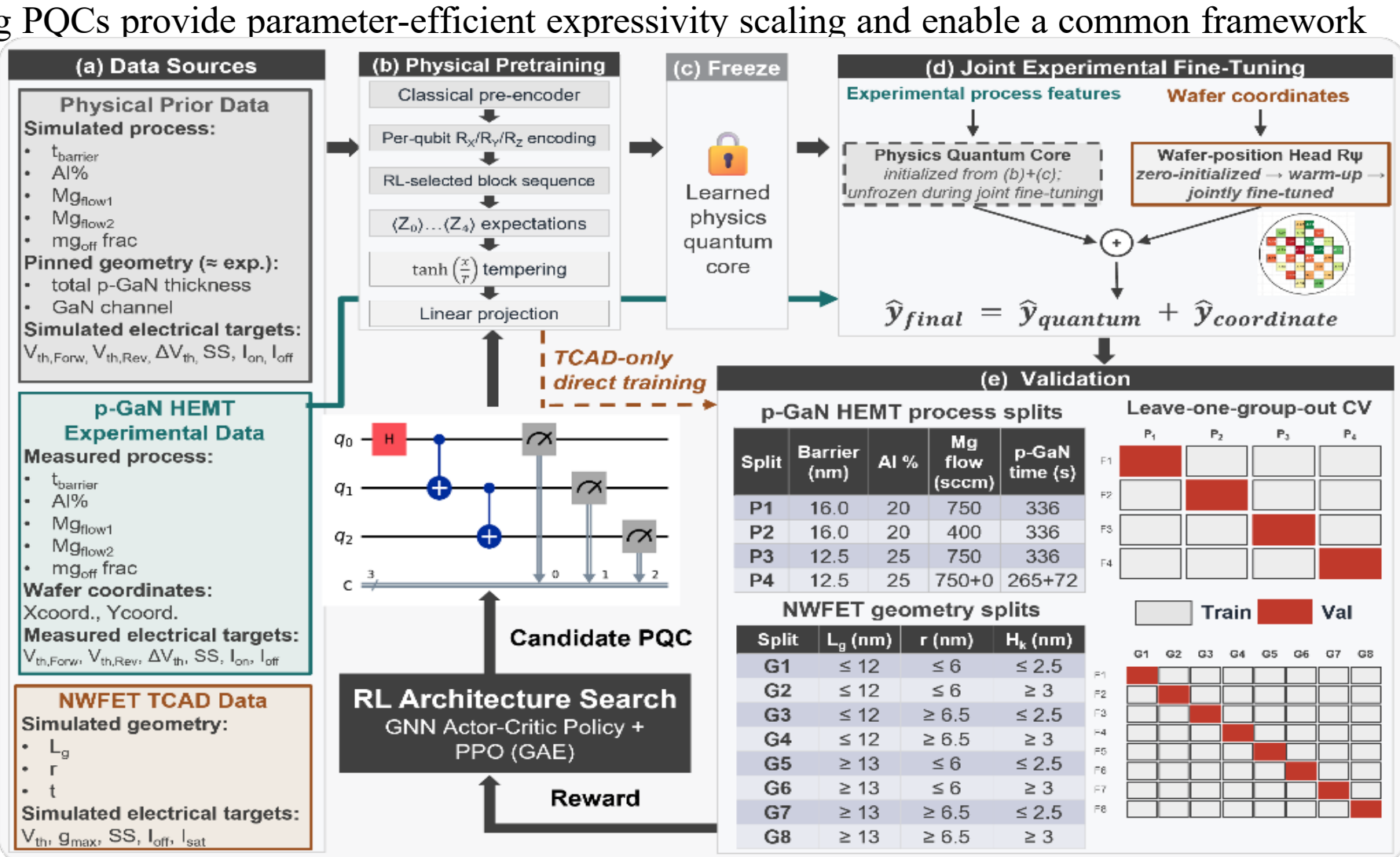

**Fig. 4.** Training and OOD evaluation workflow for RL-searched quantum surrogate models: (a) physics-prior data for pretraining and experimental data for fine-tuning; (b) simulated-data pretraining initializes the PQC physics core; (c) the pretrained core is frozen; (d) experimental fine-tuning updates the unfrozen core and wafer-position correction head; and (e) nested LOGOCV measures OOD performance and supplies the RL reward. For NWFET, the workflow uses direct TCAD-only training due to lack of available experimental data.

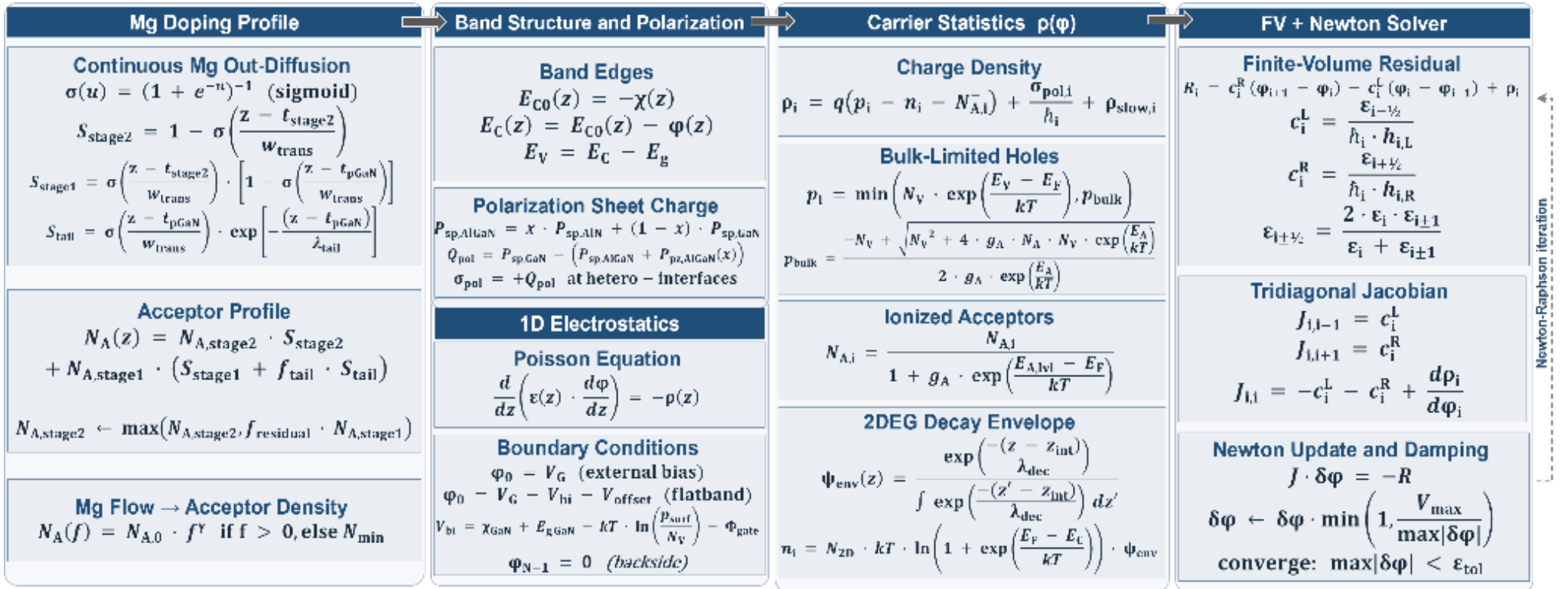

**Fig. 5.** Governing equations of the electrostatic solver for p-GaN HEMTs with considering the parameters in Table 2.

**Table 2.** Parameters used in the solver.

| Parameter | Description | Value |
|---|---|---|
| $\varepsilon_{r,GaN}$ | Relative permittivity of GaN | 9.5 |
| $\varepsilon_{r,AlGaN}$ | Relative permittivity of AlGaN | 9.5−0.5x |
| $P_{sp,GaN}$ | Spontaneous polarization of GaN | −0.034 C/m² |
| $P_{sp,AlN}$ | Spontaneous polarization of AlN | −0.090 C/m² |
| $P_{pz,AlGaN}$ | AlGaN piezoelectric polarization model | $-0.0525x+0.0282x^2$ |
| $g_A$ | Mg acceptor degeneracy factor | 4.0 |
| $E_A$ | Mg acceptor activation energy | 0.160 eV |
| $N_V$ | Valence-band effective density of states | $2.5\times10^{19}$ cm$^{-3}$ |
| $N_C$ | Conduction-band effective density of states | $2.2\times10^{18}$ cm$^{-3}$ |
| $\mu_0$ | Nominal low-field mobility prefactor | 2000 cm$^2$/(V·s) |
| $V_{DS}$ | Drain bias for low-field current | 1.0 V |
| $L_{gate}$ | Gate length | 0.8 μm |
| $E_{g,GaN}$ | GaN energy gap | 3.4 eV |
| $\chi_{GaN}$ | GaN electron affinity | 4.1 eV |
| $E_{g,AlN}$ | AlN energy gap | 6.2 eV |
| $\chi_{AlN}$ | AlN electron affinity | 0.6 eV |
| $C_{bow}$ | AlGaN bandgap bowing parameter | 1.0 eV |
| $N_{2D}$ | 2-D density of states in GaN | $8.36\times10^{13}$ cm$^{-2}$ eV$^{-1}$ |
| $\Phi_{gate}$ | Schottky gate metal work function | 4.8 eV |
| $w_{trans}$ | Doping transition width | 2.0 nm |
| $\lambda_{tail}$ | Mg out-diffusion tail length | 5.0 nm |
| $f_{tail}$ | Mg out-diffusion tail fraction | 0.02 |
| $\lambda_{decay}$ | 2DEG envelope decay length | 3.0 nm |
| $f_{active}$ | Fixed acceptor activation fraction | 0.045 |
| $N_t$ | Slow-trap sheet density (sampled) | $10^{11}$–$10^{13}$ cm$^{-2}$ |
| $E_{t,offset}$ | Trap energy offset above $E_C$ (sampled) | 0.1–0.8 eV |
| $T_s$ | Trap relaxation time constant (sampled) | $10^{-2}$–$10^{3}$ s |
| Mg coeffs | $N_A[cm^{-3}] = A(\text{Mg flow[sccm]})^B$ | $A=3.251\times10^{17}$; B=0.622 |

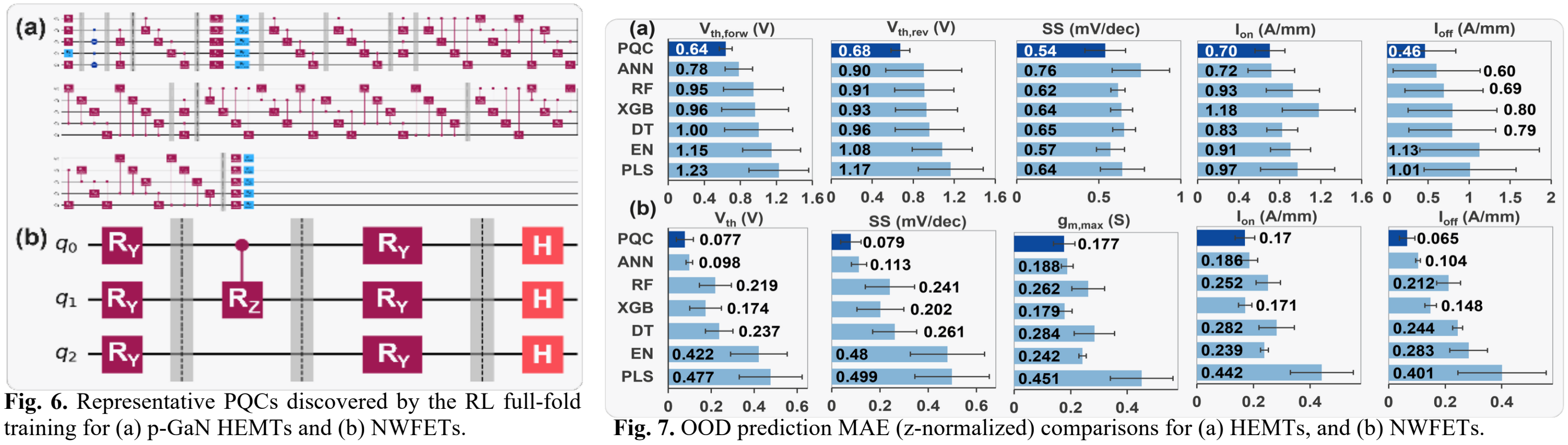

**Fig. 6.** Representative PQCs discovered by the RL full-fold training for (a) p-GaN HEMTs and (b) NWFETs.

**Fig. 7.** OOD prediction MAE (z-normalized) comparisons for (a) HEMTs, and (b) NWFETs.

**Fig. 8.** Pareto-frontier of OOD prediction error z-normalized versus trainable model parameters for (a) p-GaN HEMT and (b) NWFET.

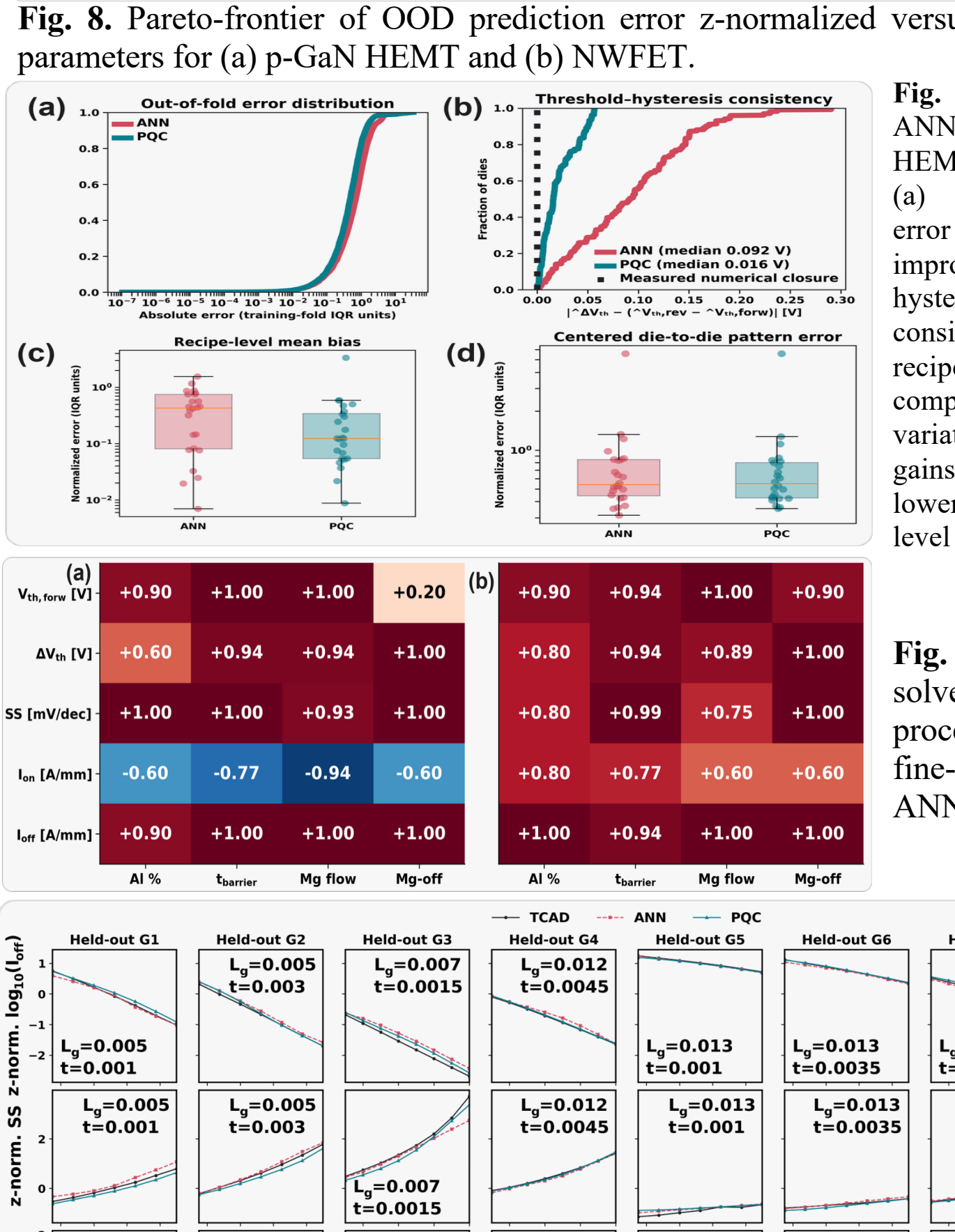

**Fig. 9.** Comparison of ANN and PQC for p-GaN HEMT OOD prediction: (a) similar out-of-fold error distributions, (b) improved threshold–hysteresis ($\Delta V_{TH}$) consistency, (c) reduced recipe-level bias, and (d) comparable die-to-die variation, indicating PQC gains mainly arise from lower systematic recipe-level bias.

**Fig. 11.** Retention of solver-learned physical process trends after fine-tuning for (a) ANN and (b) PQC.

**Table 3**. GPU speedup for dataset generation/training.

| Approach | Hardware | Runtime | Throughput | Speedup |
|---|---|---|---|---|
| **Pretraining Dataset Generation** | | | | |
| Commercial TCAD simulation | CPU | days-months | ~10-30 devices/h | 1.0x |
| Physics-based solver | CPU | ~8.4h | 1,785 devices/h | ~90x |
| PhysicsNeMo PINN | GPU | ~1.2h | 12,898 devices/h | ~650x |
| **Analytical Parameter-shift Model Training** | | | | |
| Standard Library | GPU | ~8h | 0.076 steps/s | 1.0x |
| CUDA-Q | GPU | ~0.3h | 2.17 steps/s | ~29x |

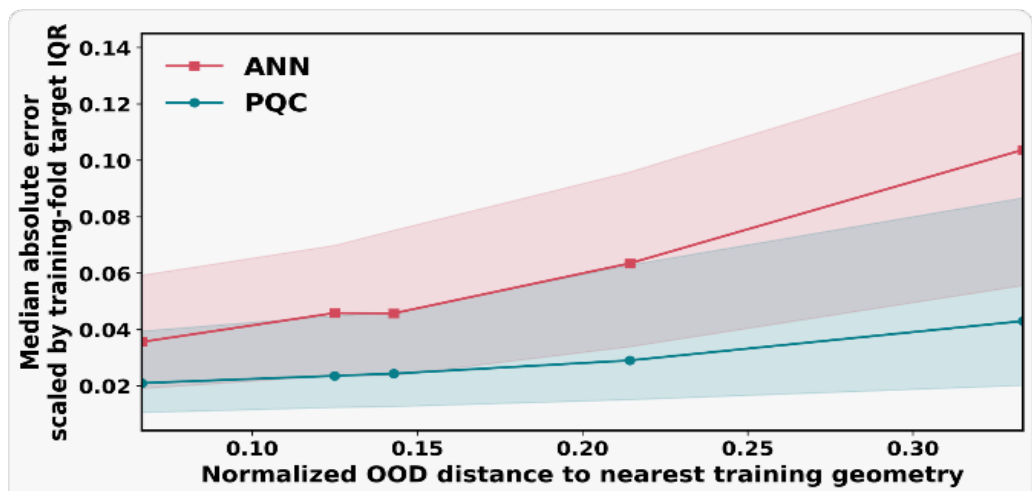

**Fig. 10.** OOD prediction error versus geometry depth for NWFETs, showing generalization to increasingly distant unseen geometries.

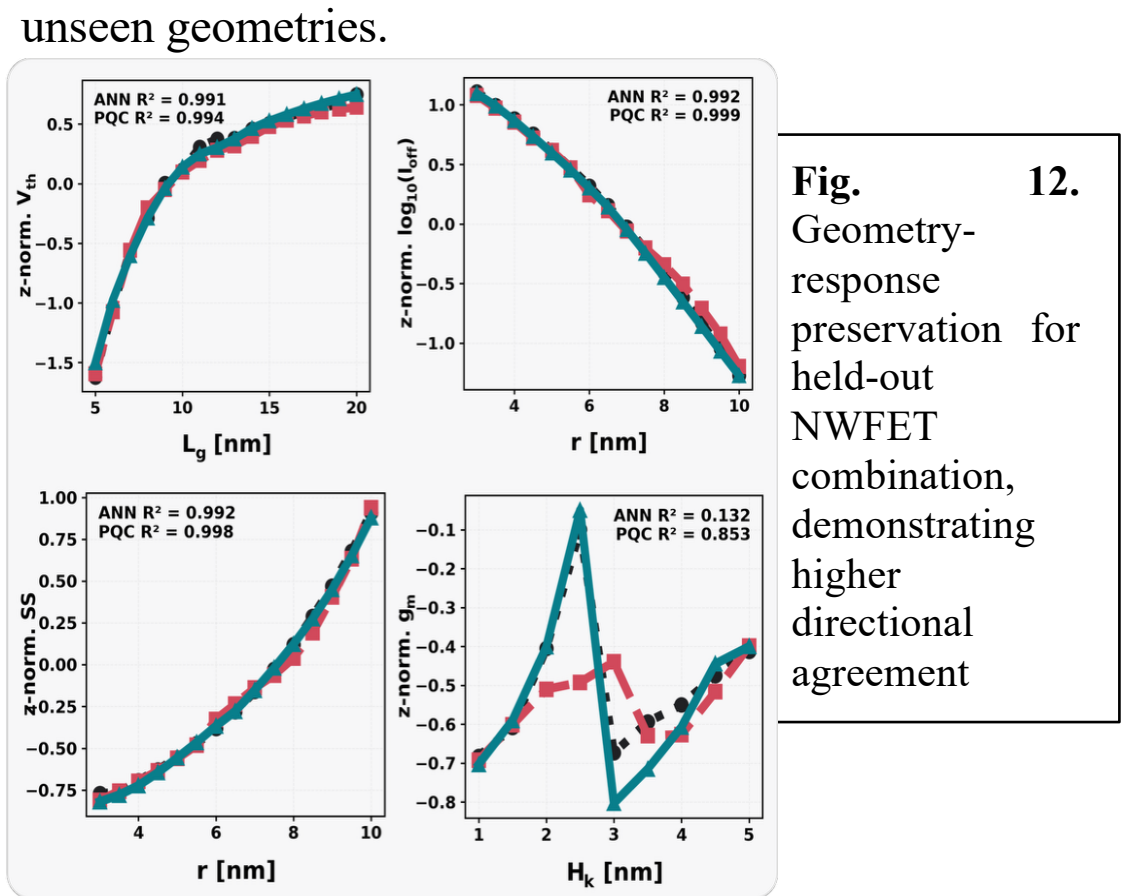

**Fig. 12.** Geometry-response preservation for held-out NWFET combination, demonstrating higher directional agreement

**Table 4.** Comparison with related works

| Ref | Dataset | Multi Target | OOD | Quantum component | Physics integration | Scalability |
|---|---|---|---|---|---|---|
| [16] | MTJ experimental + physical model | Yes | No | No | Physical equations embedded in GNN/DNN | Within MTJ/MRAM family |
| [17] | FeFET TCAD + experimental | Yes | No | No | FeFET physics features embedded in GNN | Within FeFET family; UTBSOI with fine-tuning |
| [18] | SiC/GaN TCAD + SPICE simulations | Yes | Yes | No | Electro-thermal PDEs embedded in PIGNN | Across WBG families; SiC and GaN devices |
| [19] | Si NWFET TCAD | Yes | Yes | No | Data-trained physics-informed module | No |
| [20] | GaN HEMT experimental | No | No | Quantum kernel | No | No |
| **Ours** | **p-GaN HEMT experimental + physical model; NWFET TCAD** | **Yes** | **Yes** | **RL-searched VQC** | **No device-specific equations embedded** | **Across distinct device families** |

**Fig. 13.** NWFET out-of-fold predictions (normalized) vs. nanowire radius (r) across characteristics.